 \documentclass[pmlr,twocolumn]{jmlr} 

\usepackage{booktabs}
\usepackage[load-configurations=version-1]{siunitx} 
\usepackage{multirow}

\theorembodyfont{\upshape}
\theoremheaderfont{\scshape}
\theorempostheader{:}
\theoremsep{\newline}

\title[Spatial-Temporal Convolutional Network for Spread Prediction of COVID-19]{Spatial-Temporal Convolutional Network for Spread Prediction of COVID-19}

\author{
\Name{Ravid Shwartz-Ziv}
\Name{Itamar Ben Ari}
\and
\Name{Amitai  Armon}\\
\Email{\{ravid.ziv,\ itamar.ben-ari,\ amitai.armon\} @intel.com}\\
\addr {Advanced Analytics, Intel }
}

\begin{document}

\maketitle

\begin{abstract}
In this work we present a spatial-temporal convolutional neural network for predicting future COVID-19 related symptoms severity among a population, per region, given its past reported symptoms. This can help approximate the number of future Covid-19 patients in  each region, thus enabling a faster response, e.g., preparing the local hospital or declaring a local lockdown where necessary. Our model is based on a national symptom survey distributed in Israel and can predict symptoms severity for different regions daily. The model includes two main parts - (1) learned region-based survey responders profiles used for aggregating questionnaires data into features (2) Spatial-Temporal 3D convolutional neural network which uses the above features to predict symptoms progression.
\end{abstract}
\begin{keywords}
Covid-19, Convolutional Neural Networks, Spatial-Temporal prediction
\end{keywords}
\section{Introduction}
One major factor contributing to the spread of the COVID-19 is the large number of undiagnosed infected individuals. These individuals increase the spread of the virus and delay the response of public health authorities, resulting in an explosion in the number of cases.

The spread of the disease presents many challenges to policy-makers across the world.  However, a fundamental resource for effective decisions is accurate information about the spread of the virus, including the prediction of its future spread levels. The virus tends to spread in regioanl clusters, and early detection of these clusters is critical for slowing down the virus' spread.

Recently, the authors of \citep{covid_framwork} proposed to use daily population-wide surveys that assess the development of symptoms caused by the virus as a tool for identifying such clusters. Here, we present a region-based model for symptom severity prediction based on a national symptoms survey distributed in Israel. Our model has two components: (1) learning region-based survey responders profiles used for aggregating questionnaires data into features and (2) a 3D convolutional neural network for predicting future symptoms severity in each region based on these features. The model considers past symptom levels in a specified time-window and can extrapolate to the future, even for non-reported areas, at a high geographical resolution.  

While most related studies are based on clinical characteristics of COVID-19 hospitalized patients \cite{covid19_survey}, our model uses population-wide daily short surveys. This enables providing daily predictions with a much higher geographical resolution and even extrapolate them for non-reported areas.  

\section{Data}
The online questionnaire from \cite{covid_framwork} contains daily information of reported symptoms across the entire Israeli population. This survey was posted online on March $14^{th} \ 2020$, and participants were asked to fill it daily. The survey is filled anonymously and was filled out $\approx 500000$ times until May $21^{th} \ 2020$.

The survey includes questions on respondents' age, gender, geographic location, isolation status, smoking habits, prior medical conditions, body temperature measurement, and self-reporting symptoms such as cough, fatigue, myalgia (muscle pain), shortness of breath, rhinorrhea or nasal congestion, diarrhea, headache, chills, and confusion. These symptoms and questions were chosen by medical professionals, based on symptoms described as prevalent in patients with COVID-19 infection and symptoms.

To analyze additional features that may affect the prevalence of symptoms in the population, we also utilized social data from the Israel Central Bureau of Statistics, such as regional demographic characteristics and socio-economic levels. 

\section{The Proposed Method}
Following \cite{covid_framwork},  we defined the Symptoms Ratio Average (SRA) for each participant as the number of reported symptoms divided by the number of symptoms in our predefined list. Our task is to predict the future SRA for each location, given the past questionnaires data. To fully utilize the features' information, we introduce a novel classification framework based on two stages - (i) pre-processing of the data to create 'good' geographical features and (ii) using these spatial-temporal features to predict the SRA for each location.

For the pre-processing part, we extract new features based on learned clusters of the original features. Then, we build a $3D$ representation of the new features for prediction. This representation is generated by stacking questionnaires' values, which preserves geographical information of the questionnaires' locations. Finally, we use $3D$ convolutional layers to aggregate both the geographic and the temporal data for predicting the SRA per location.

Using our model, given the input for the last $n$ days, our goal is to predict the SRA values for all the locations in the $k-th$ day in the future. Our focus in this work was predicting for the next day ($k=1$).

\subsection{Features extractor by learned profiles}
Mapping questionnaires data into a single feature vector can be done naively by averaging questionnaires values per survey question. Since different population profiles (e.g. old vs. young population) express symptoms differently, we want to represent each population profile with its own feature vector. The profiles were defined by a Gaussian mixture model computed by an Expectation-Maximization  (EM)  algorithm \citep{moon1996expectation} over the questionnaires values vector. Each responder is mapped to the nearest GMM cluster. We then averaged questionnaires mapped to the same cluster, which resulted in a feature vector per cluster/profile. We concatenate the feature vectors of all clusters into a single feature vector. This operation is done per geographic location and time window (see section \ref{sec:rep_ques} below).      

\subsection{4D Representation of the questionnaires' values}
\label{sec:rep_ques}
We next build a $4D$ representation of the questionnaires' values. The first step is transforming all the features derived from the pre-processing for each date into a fixed $2D$ grid, via binning based on the spatial location where the questionnaires where filled. For a point with no data, we interpolate it with values from the closest geographical regions. This step preserves the questionnaires' spatial information and allows us to use it in the prediction step. Then, we stack different features on top of each other as channels in an image. This multi-channel image is a complete representation of the questionnaires' values for a specific point in time. In the last step, we stack different dates to create a $4D$ tensor representation, preserving the temporal information. 
\subsection{Convolutional Encoder-Decoder model for prediction}
To consider both the geographical and temporal information from previous dates, we look at the extracted $4D$ features above as frames in a video where the goal is to predict only one channel: the SRA of the next frames (future dates).  Following the success of $3D$ CNNs for video recognition \citep{arunnehru2018human}, we employ  $3D$ Spatio-temporal convolutional layers. These layers, stacked together in an encoder network, transfer the input to a latent space. Then, through upsampling by deconvolutional layers \cite{zeiler2010deconvolutional} we produce the $2D$ prediction of the SRA for the target date. The last layer of the decoder outputs two values for each pixel. These values take as the SRA mean and standard deviation in the location represented by that pixel, where we train by optimizing the maximum log-likelihood of this Gaussian with the true SRA levels.  

Note that since our algorithm considers geographical information, we can make predictions even for an area without data or with very little data. This feature is critical sine there are often sub-populations that are under-represented in the survey.

\subsection{Features size and resolution}
Using a naive approach, we can feed our network with the data of the whole country at once. Our model's input would be all the locations in our data. However, in the case we do not have enough data for a high-resolution prediction, we can use instead patches of smaller size. Using this approach, our model needs to provide a  prediction only for local areas, which may be a much easier task. The cropping operation of the full image to the different patches can either be done arbitrarily or by location (creating a patch around each region of interest).

Another important hyperparameter of the network is the resolution of the input and output. We have the classical bias-variance trade-off \citep{kohavi1996bias}) in learning from finite data. If our resolution is too high, we do not have enough examples to learn each pixel, while for a too low resolution, our prediction is not specific enough. 

\section{Experiment Results}
\subsection{Experimental setup}
For the encoder, we use two $3D$ convolutional layers following a dense layer with a size of $128$. For the decoder, we use two $3D$ deconvolution layers following one $2D$ deconvolution layer for the output. The resolution for both the input and the output is $77 \times 29$ for the whole image, and the patch size is $10 \times 10$.

We performed  the binning of the data for each day. We split it to windows of size four days each, where the first three days for each window are the input, and the SRA of the fourth day is the label. Finally, we took the last two weeks of the data for the test set and did cross-validation on the rest of the windows. We calculated the $R^2$ for each validation set during the cross-validation and took the mean overall the cross validation's iterations to optimize the hyperparameters. All results
are the mean of the test scores over the different partitions, and the standard error of the mean is reported. The optimizer that we use is Adam with a learning rate of $1e^{-4}$, where the loss function is a Maximum Likelihood on a truncated Gaussian. 

For comparison, we use a baseline model, a network with two fully connected layers which operates on the whole image for a specific date. This baseline does not consider neither the temporal nor the spatial correlations in our data.

\subsection{Results}
\subsubsection{The different components }
Through the setup above, we compare our model to several models: (1) the baseline model, (2) a model that was trained on the full images, and (3) a  model that was trained on the raw features without the learned profiles.
In table \ref{ta:results}, we compare the performance for all the models (both $R^2$ and Spearman Correlation (SC)).  Our model outperformed all the other models, where the baseline FC model was the worst. Although each component of the final model is necessary for the final results, it seems that the most important one is extracting  the features extracted by the learned profiles (the raw features model had bad performance).

\begin{table}
  \centering
\label{ta:results}
 \begin{tabular}{||c | c | c||} 
 \hline
 Model & R2 & SC \\ [0.5ex] 
 \hline
 \multirow{2}{*}{Base model} &0.129 &0.11\\ & \footnotesize{$(\pm0.028)$} & \footnotesize{$(\pm0.039)$} \\ 
    
 \hline
  \multirow{2}{*}{3D Conv - Raw Features} &0.314 &0.21\\ & \footnotesize{$(\pm0.013)$} & \footnotesize{$(\pm0.020)$} \\ 

  \hline
  \multirow{2}{*}{ 3D Conv - Full Images} &0.39 &0.26\\ & \footnotesize{$(\pm0.026)$} & \footnotesize{$(\pm0.015)$} \\ 

 \hline
  \multirow{2}{*}{\textbf{3D Conv - Full model}} &0.501 &0.31\\ & \footnotesize{$(\pm0.012)$} & \footnotesize{$(\pm0.009)$} \\ 

  \hline
 
\end{tabular}
  \caption{The different models' components} \label{tab:sometab}
\end{table}

\subsubsection{The effect of the resolution}
As mentioned before, we need to pre-define our input resolution. Each questionnaire in our data has a specific physical location. Originally, the resolution of this location can get to tens of meters. However, because we do not have enough reported questionnaires for each location, we must aggregate data from close locations. There are several options to deal with this problem, and in our model we decided to do binning to pre-defined bins.
The size of the bins is a crucial hyper-parameter. Increasing the bin's size will aggregate more data in each bin but lose the spatial information of their location. 

\begin{figure}[h]
    \centering
    \includegraphics[width=0.8\columnwidth]{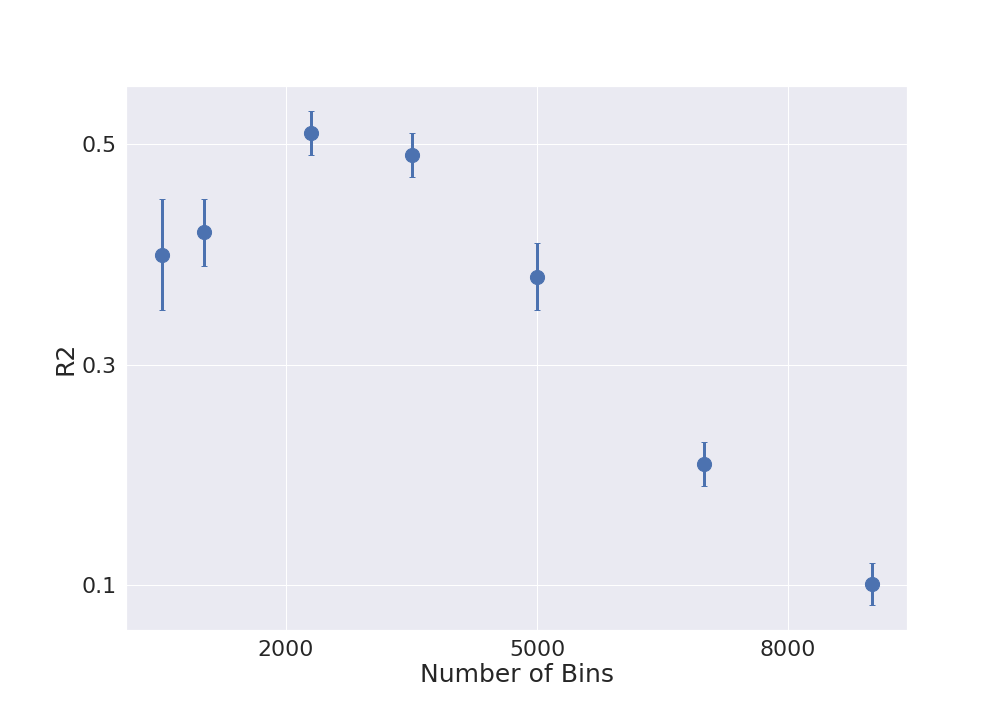}
    \caption{$R^2$ for different resolutions}
    \label{fig:resolution}
\end{figure}

In figure \ref{fig:resolution}, we present the performance of our model ($R^2$ score) as a function of the number of bins in the input (the resolution). We can see that there is a peak around $3000$ bins, which represents the best point in the trade-off between enough spatial information and enough data.  

\section{Discussion}
 We developed a Spatial-Temporal convolutional network for predicting the future geographical spread of COVID-19 related symptoms. This model, based on data obtained by a quick and easy survey, allows us to give a prediction regularly with a much higher geographical resolution. Through the use of $3D$ convolutions, our model is capable of learning spatiotemporal features, and it outperforms other models. We investigated the different parts of our model,  their effect on the performance, and the resolution trade-off.

\bibliography{main}

\end{document}